\documentclass{article}
\usepackage{spconf,amsmath,graphicx}
\usepackage{array}
\usepackage{booktabs}
\usepackage{multirow}
\usepackage{marvosym}

\usepackage{enumitem}
\setlist{nosep, leftmargin=14pt}

\usepackage{mwe} % to get dummy images

\title{Clinical Inspired MRI Lesion Segmentation}
\name{Lijun Yan\textsuperscript{$\star$} \qquad Churan Wang\textsuperscript{$\dagger$ \Letter} \qquad Fangwei Zhong\textsuperscript{$\ddagger$} \qquad Yizhou Wang\textsuperscript{$\dagger$}}
\address{
\textsuperscript{\Letter} Correspondence to: \textit{churanwang@pku.edu.cn}\\
$^{\star}$School of Software and Microelectronics, Peking University, China.\\
$^{\dagger}$Center on Frontiers of Computing Studies, School of Computer Science,\\
                 Nat'l Eng. Research Center of Visual Technology, Peking University, China.\\
$^{\ddagger}$School of Artificial Intelligence, Beijing Normal University, China.
}
\begin{document}
%\ninept
%
\maketitle
\begin{abstract}
% Magnetic resonance imaging (MRI) is a potent diagnostic tool for detecting and localizing pathological tissues in various diseases. However, different MRI sequences may have different contrast mechanisms and sensitivities for different types of lesions, which pose challenges for accurate and consistent lesion segmentation. In this paper, we propose a novel method for MRI lesion segmentation that is inspired by clinical experience. Our method leverages the idea of residual learning, which aims to learn the difference between post contrast-enhanced T1-weighted (post) and pre contrast-enhanced (pre) sequences, which are commonly used by radiologists to locate lesions. Specifically, we iteratively and adaptively fuse features from pre- and post-contrast images at multiple resolutions, using dynamic weights to achieve optimal fusion and address diverse lesion enhancement patterns. Our method achieves state-of-the-art performances on the public BraTS2023 dataset for brain tumor segmentation and our in-house breast MRI dataset for breast lesion segmentation. Our method is clinically inspired and has the potential to facilitate lesion segmentation in various applications.
Magnetic resonance imaging (MRI) is a potent diagnostic tool for detecting pathological tissues in various diseases. 
% It features numerous different sequences that reflect various internal structures of the human body. 
Different MRI sequences have different contrast mechanisms and sensitivities for different types of lesions, which pose challenges to accurate and consistent lesion segmentation.
In clinical practice, radiologists commonly use the sub-sequence feature, i.e. the difference between post contrast-enhanced T1-weighted (post) and pre-contrast-enhanced (pre) sequences, to locate lesions.
Inspired by this, we propose a residual fusion method to learn subsequence representation for MRI lesion segmentation. 
Specifically, we iteratively and adaptively fuse features from pre- and post-contrast sequences at multiple resolutions, using dynamic weights to achieve optimal fusion and address diverse lesion enhancement patterns. Our method achieves state-of-the-art performances on BraTS2023 dataset for brain tumor segmentation and our in-house breast MRI dataset for breast lesion segmentation. Our method is clinically inspired and has the potential to facilitate lesion segmentation in various applications.
\end{abstract}

\vspace{-0.3cm}
\section{Introduction}
\vspace{-0.3cm}
\label{sec:intro}
Magnetic Resonance Imaging (MRI), a non-invasive and radiation-free 3D medical imaging technique, plays a crucial role in screening~\cite{mriback5}, diagnosing, and treating various diseases due to its high sensitivity and detailed anatomical visualization. The initial MRI scanning involves T1-weighted sequences, known as pre-contrast sequences. Then, a contrast agent, typically Gadolinium (Gd), is administered, followed by multiple scans of T1 sequences, creating post-contrast sequences~\cite{bougias2022breast}. Post-contrast sequences highlight structures like blood vessels, inflammation, and lesions with elevated signal intensity. Comparing pre-contrast and post-contrast sequences enables doctors to identify potential lesions.

\begin{figure}[t!]
\includegraphics[width=8.5cm]{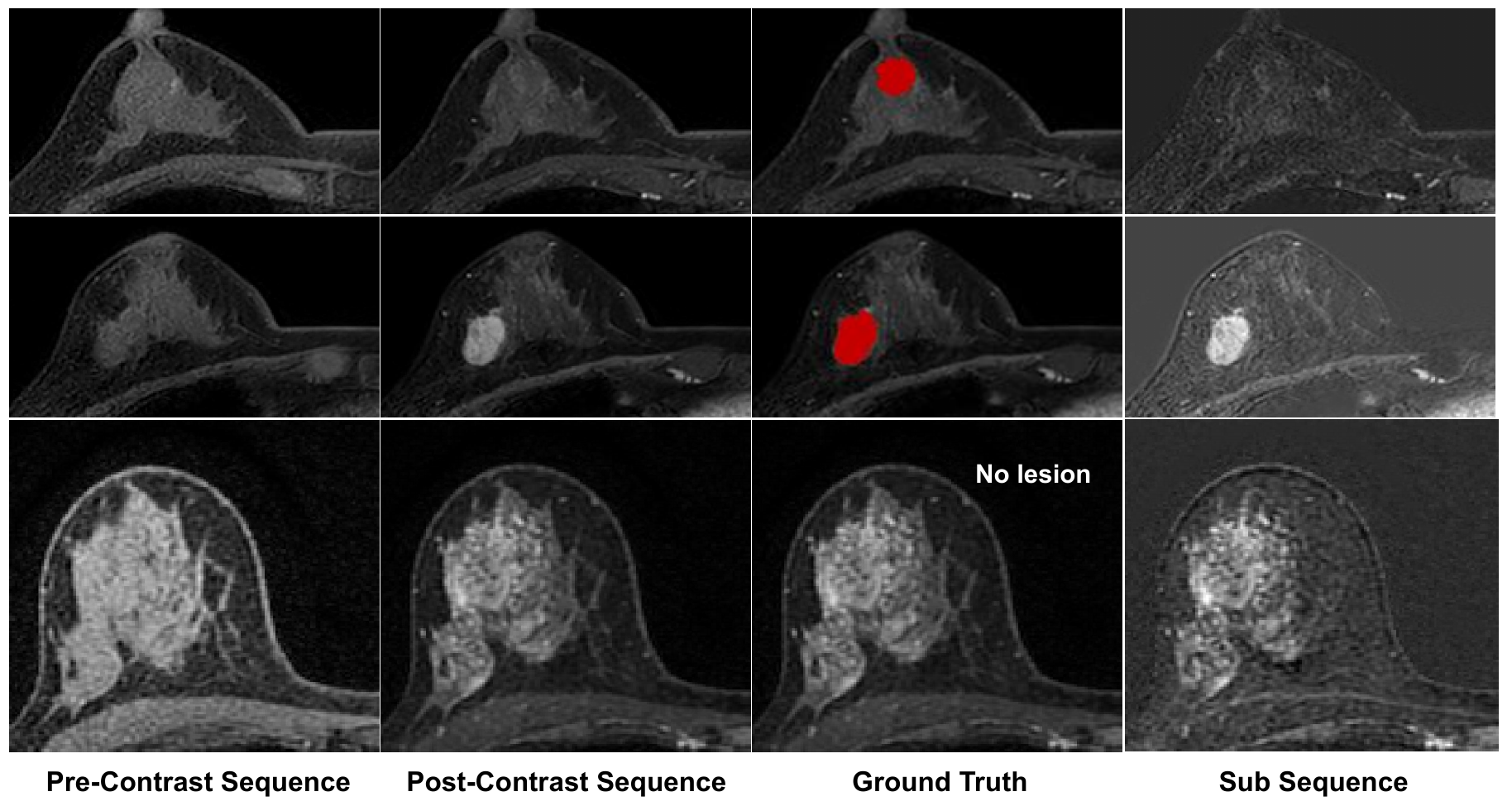}
\vspace{-0.4cm}
\caption{
% Overview of enhancement patterns. The four columns represent pre-contrast images, post-contrast images, the lesions drawn by radiologists, and the subtraction images of post- and pre-contrast images (SUB).
% The patient in the first row is diagnosed with fibroadenoma, showing a low degree of enhancement in the post-contrast image. Since neither the lesion nor the gland enhances, the SUB image contains little useful information.
% The second row represents a typical mass, characterized by a low signal in the pre-contrast image and high enhancement in the post-contrast image.
% In the third row, the patient exhibits a well-vascularized glandular tissue, displaying high signal intensity in both pre- and post-contrast images, without any lesion, resulting in a visibly enhanced portion in the SUB image. Hence, Relying solely on the SUB image may lead to false positives.
Overview of enhancement patterns. The four columns represent pre-contrast images, post-contrast images, the lesions drawn by radiologists, and the subtraction images (SUB). The three rows represent fibroadenoma, typical mass, and well-vascularized glandular tissue respectively.
} \label{intro}
\vspace{-0.4cm}
\end{figure}

Distinct enhancement patterns are observed in different tissues~\cite{mann2019breast,smirniotopoulos2007patterns}. Tissues with abundant blood supply, such as tumors and inflammation, often exhibit high signal intensity in enhanced images. Conversely, structures like cysts and calcification show no enhancement. Furthermore, certain healthy tissues may undergo enhancement; for instance, normal breast fibroglandular tissue may appear highly enhanced in post-contrast sequence due to its rich blood supply (Fig.~\ref{intro} illustrates these scenarios). Given the diversity in enhancement patterns, clinical diagnosis requires radiologists to analyze both pre-contrast and post-contrast sequences, because depending solely on post-contrast sequences for diagnosis is one-sided and error-prone and a comprehensive diagnosis must integrating pre-contrast information.

An intuitive idea to achieve the fusion of pre-contrast and post-contrast MRI sequences is to directly utilize pre-contrast and post-contrast images and design attention-based mechanisms\cite{galli2021pipelined,wang2023a2fseg,han2023explainable,zhang2023multi,wang2023breast,hatamizadeh2022unetr,rahman2023ambiguous}. Meanwhile, some methods focus on learning the enhancement process and extracting corresponding lesion information~\cite{lv2023diffusion,chung2023deep}. However, direct fusion approaches may not fully utilize the distinct and complementary information each sequence offers. Moreover, they may not adequately address the variations in image characteristics, such as noise levels and contrast differences, potentially impacting the accuracy of lesion characterization. 

% An intuitive idea to achieve the fusion of pre-contrast and post-contrast MRI sequences is to directly utilize pre-contrast and post-contrast images and design attention-based mechanisms\cite{galli2021pipelined}. For example, Zhong \textit{et al.}~\cite{zhong2023simple} proposed a weakly supervised breast cancer segmentation method based on landmark annotations, aiming to learn more representative features of lesions, meanwhile, some approches~\cite{wang2023a2fseg,han2023explainable,zhang2023multi,wang2023breast,zhou2023cocoattention} focused on the direct fusion of image features and some new backbones of lesion segmentation were proposed\cite{hatamizadeh2022unetr,rahman2023ambiguous}. Some methods focus on learning the enhancement process. Lv \textit{et al.}~\cite{lv2023diffusion} used a diffusion model to learn and simulate kinetic information during the enhancement process, and the acquired information was then utilized for lesion segmentation. Similarly, Chung \textit{et al.}~\cite{chung2023deep} also attempted to simulate the post-contrast sequences. However, direct fusion approaches may not fully capitalize on the distinct and complementary information each sequence offers. Moreover, they may not adequately address the variations in image characteristics, such as noise levels and contrast differences, potentially impacting the accuracy of lesion characterization. 

To solve this problem, we focus on how to naturally explore the relationship between pre-contrast and post-contrast images. The residual concept aligns more closely with the diagnostic process, where radiologists often look for changes between sequences to make informed decisions. Inspired by residual learning \cite{He_2016_CVPR} and  counterfactual generation\cite{churanmiccai,churantip}, we propose a residual learning framework. Our approach focuses on learning the differences or ‘residuals’ between pre-contrast and post-contrast images. We involve integrating pre-contrast images into the segmentation encoder at each encoding step, enabling the network to focus on learning enhancement patterns. By doing so, our network can more effectively highlight the changes due to contrast enhancement, which is critical for accurate lesion detection and segmentation. Hence, our method not only improves the integration of MRI sequences but also enhances the clinical relevance of the segmentation results. In general, our work contributes in the following ways: 
\begin{itemize}
    \item We introduce a novel neural network architecture that seamlessly incorporates pre-contrast information into post-contrast images, contributing to enhanced lesion diagnosis and segmentation, with only a minimal increase in the parameters of the backbone network.
    \item Aligning with the diagnostic practices of radiologists, our method leverages more comprehensive information, enhancing lesion diagnosis and segmentation efficacy.
    % leading to superior results in lesion diagnosis and segmentation.
    \item Our model achieves state-of-the-art performance in brain tumor and breast lesion segmentation tasks.
\end{itemize}

\vspace{-0.2cm}
\section{Method}
\vspace{-0.2cm}
% In our study, we present a model that effectively integrates information from both pre-contrast T1-weighted (pre-T1) and post-contrast T1-weighted (post-T1) images, aimed at enhancing tumor medical image analysis. The schematic of our network architecture is illustrated in Fig.~\ref{arch}.

% Our method involves extracting image features from both pre- and post-contrast images using shared neural networks within the encoding blocks. These blocks utilize convolutional architectures to capture both local and global features.
% Acknowledging that post-T1 images typically exhibit enhanced brightness for most lesions, we designate the post-T1 image as the primary branch of our model, termed the main branch. The pre-T1 image serves as an auxiliary branch, offering complementary contrasting information. This strategic design aims to leverage the enhanced information inherent in post-T1 images for more effective tumor analysis.

\begin{figure*}[t!]
\centering
\includegraphics[width=\textwidth]{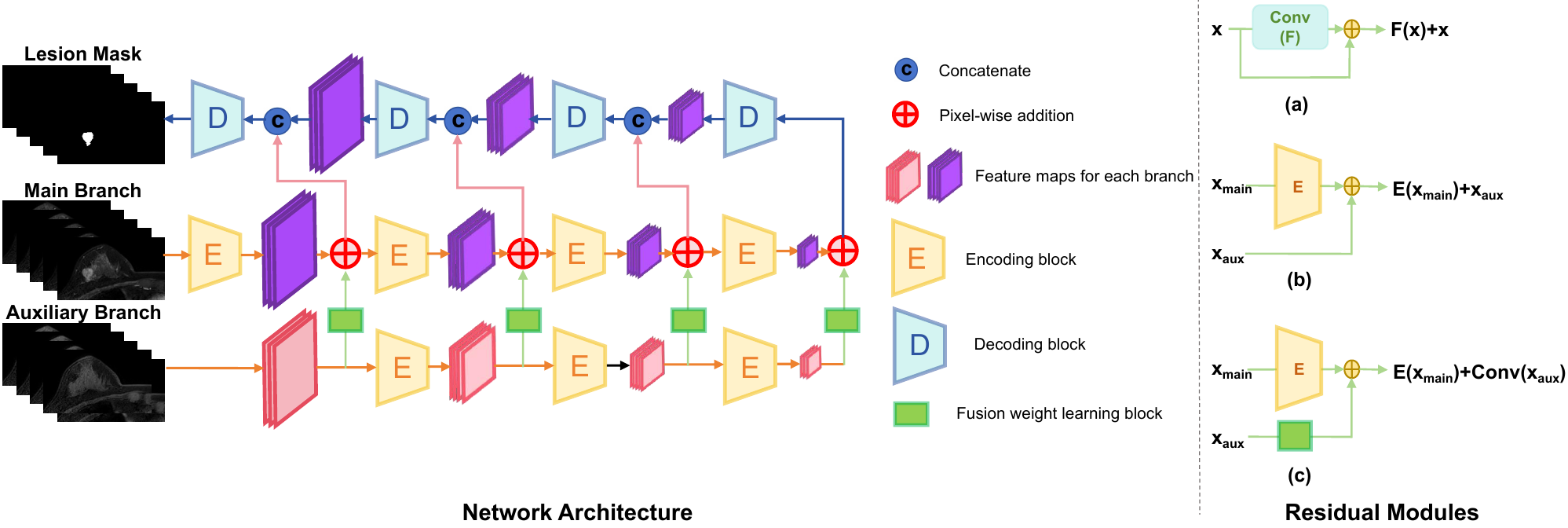}
\vspace{-0.8cm}
\caption{\textbf{Left}: The architecture of our network, featuring a main branch and an auxiliary branch, each comprising encoding blocks, as well as fusion weight learning blocks. \textbf{Right}: residual modules of different versions. (a) is the original version from ResNet, (b) is our modified version which incorporates two branches, and (c) is our final residual module which adds a fusion weight learning block into (b).} \label{arch}
\vspace{-0.3cm}
\end{figure*}

% To further enhance the model's capability in capturing diverse image details, we implement a multi-scale feature fusion mechanism, which learns residuals of pre- and post-contrast image features at various scales within the neural network. By incorporating multi-scale feature fusion, our model can effectively grasp intricate information across different scales present in the images, thereby contributing to improved performance in lesion exploration.
Our study proposes a model integrating pre- and post-contrast T1-weighted images to enhance tumor analysis, with the network architecture shown in Fig.~\ref{arch}. We extract features from both images using shared neural networks in encoding blocks, capturing local and global features via convolutional architectures. Post-T1 images, with enhanced brightness for lesions, form the main branch, while pre-T1 images serve as an auxiliary branch. Additionally, a multi-scale feature fusion mechanism is implemented to learn residuals of image features at various scales, enabling the model to grasp intricate details across scales and improve lesion exploration performance.

% \begin{figure*}[t!]
% \centering
% \includegraphics[width=\textwidth]{arch_2.pdf}
% \caption{Architecture of our proposed network, featuring a main branch and an auxiliary branch, each comprising encoding blocks, as well as fusion weight learning blocks. Encoded features from the main branch and auxiliary branch are interconnected through pixel-wise addition operations at multi-scale. Finally, decoding blocks and skip connections are utilized to generate the segmentation mask.} \label{arch}
% % \vspace{0.3cm}
% \end{figure*}

\vspace{-0.3cm}
\subsection{Residual Module \& Multi-scale Feature Aggregation}
\vspace{-0.2cm}
%Derived from the ResNet architecture~\cite{He_2016_CVPR}, we incorporate skip connections into the encoding process, departing from the traditional Res-UNet~\cite{zhang2018road} approach where input features directly link to post-convolutional features. Our innovation involves connecting feature maps from the auxiliary branch to the post-convolutional features of the main branch, fostering a feature fusion step between the main and auxiliary branches.

Derived from the ResNet architecture~\cite{He_2016_CVPR}, we incorporate skip connections into the encoding process, departing from the traditional Res-UNet~\cite{zhang2018road} approach where input features directly link to post-convolutional features. Our innovation involves residual learning of the main branch and the auxiliary branch, fostering a feature fusion step between the main and auxiliary branches.

% \begin{figure}[t!]
% \centering
% \includegraphics[width=7cm]{residual.pdf}
% \caption{Residual modules of different versions. (a) is the original version from ResNet, (b) is our modified version which incorporates two branches. (c) is our final residual module which adds a fusion weight learning block into (b).} \label{residual}
% % \vspace{0.3cm}
% \end{figure}

%加入一段对residual的说明
Inspired by residual learning in ResNet, we propose a method to learn the residual between two branches. Fig.~\ref{arch} illustrates the residual module in ResNet and the modified residual module we derived. Fig.~\ref{arch}(a) shows the original residual module from the ResNet paper, which improves training efficiency and stability by learning the difference (i.e. residual F(x)) between the input x and the desired output H(x)
% , as presented in \eqref{res}.
%  \begin{equation}
%     H(x)=F(x)+x \label{res}
% \end{equation}
Considering that the MRI images before and after enhancement theoretically only have significant differences in the lesion areas and have very similar structures, we modify the residual module, which is shown in Fig.~\ref{arch}(b). We set the desired output as $H(x)=E(x_{main})+x_{aux}$, which enables the network to more efficiently learn lesion-related features through the residuals between the two branches.

Throughout the multiple down-sampling stages of the encoder-decoder architecture, we repeatedly use the modified residual module mentioned above, ensuring the integration of both the main and auxiliary branches across different resolutions. This approach aims to enhance the network's capacity to capture intricate details and contextual information at diverse levels.  By linking feature maps from the auxiliary branch to the main branch, our model can leverage both the detailed information learned by the main branch and the contrasting information provided by the auxiliary branch.

% Throughout the encoder-decoder architecture's multiple down-sampling stages, we perform feature fusion at various scales, ensuring integration of both main and auxiliary branches across different resolutions, which is achieved by learning residuals of two branches. This approach aims to enhance the network's capacity to capture intricate details and contextual information at diverse levels.
% By linking feature maps from the auxiliary branch to the main branch post-convolutional operations, our model can leverage both the detailed information learned by the main branch and the contrasting information provided by the auxiliary branch.

This feature fusion strategy not only maintains the network's high-resolution information capture capability but also enriches feature representations with complementary details from both branches. The resulting network architecture is well-suited for medical image analysis tasks, particularly in scenarios where accurate diagnosis and segmentation require the incorporation of intricate details and subtle contrasts.

\vspace{-0.4cm}
\subsection{Dynamic Learning of Auxiliary Branch Weights}
\vspace{-0.2cm}
The fusion of features from both branches involves a direct addition operation. However, recognizing the need for different weights for features of the two branches, we introduce a block designed to dynamically learn the weights of the auxiliary branch before the addition operation. Specifically, we employ a 1*1*1 convolutional layer to dynamically learn these weights. The schematic of the final residual module is shown in Fig.~\ref{arch}(c). This approach introduces a minimal number of parameters, ensuring efficiency, while also allowing for adaptable weight adjustments for the auxiliary branch. 

% Our experimental results demonstrate that the introduced feature fusion mechanism significantly contributes to the overall performance improvement of the backbone architecture. It is crucial to highlight that, except for the component associated with dynamic weight learning, no additional parameters are introduced compared to the backbone. This ensures that there is no significant memory burden during training. Furthermore, our proposed modules exhibit versatility by seamlessly accommodating various backbones.

% \begin{table*}[t!]
% \centering
% \caption{Comparison among recent methods on our in-house breast MRI dataset.}\label{result_breast}
% \begin{tabular}{p{3cm}|p{2cm}<{\centering}|p{2cm}<{\centering}}
% \hline
% \textbf{Methodology} & \textbf{DSC} & \textbf{Recall} \\
% \hline
% Res-UNet-Single~\cite{zhang2018road} & 78.10\% & 74.47\%\\ 
% \hline
% Res-UNet-Concat~\cite{zhang2018road} &78.21\% & 77.86\%  \\ 
% \hline
% A2FNet~\cite{wang2023a2fseg} & 80.01\%&82.42\% \\ 
% \hline
% TSF-Seq2Seq~\cite{han2023explainable} & 80.22\%& 80.63\%\\ 
% \hline
% LiTS~\cite{zhang2023multi} &79.99\%& 79.50\%   \\ 
% \hline
% Ours &\textbf{82.55\%} &82.05\% \\ 
% \hline
% \end{tabular}
% \vspace{-0.5cm}
% \end{table*}

\begin{table}[t!]
\centering
\caption{Comparison among recent methods on our in-house breast MRI dataset.\label{result_breast}}
\begin{tabular}{>{\centering\arraybackslash}p{4cm} % 使用booktabs的列设置
          >{\centering\arraybackslash}p{1.5cm} % 同上
          >{\centering\arraybackslash}p{1.5cm}} % 同上
\toprule % 使用booktabs的toprule
\textbf{Methodology} & \textbf{DSC} & \textbf{Recall} \\
\midrule % 使用booktabs的midrule
ResUNet\cite{zhang2018road} & 78.10\% & 74.47\% \\ 
% Res-UNet-Concat\cite{zhang2018road} & 78.21\% & 77.86\% \\ 
A2FNet\cite{wang2023a2fseg} & 80.01\% & 79.54\% \\ 
TSF-Seq2Seq\cite{han2023explainable} & 80.22\% & 80.63\% \\ 
LiTS\cite{zhang2023multi} & 79.99\% & 79.50\% \\ 
Ours & \textbf{82.55\%} & \textbf{82.05\%} \\ 
\bottomrule % 使用booktabs的bottomrule
\end{tabular}
\vspace{-0.3cm}
% 移除\vspace命令，因为它可能会影响表格的布局
\end{table}

\begin{figure}[t!]
\includegraphics[width=8.5cm]{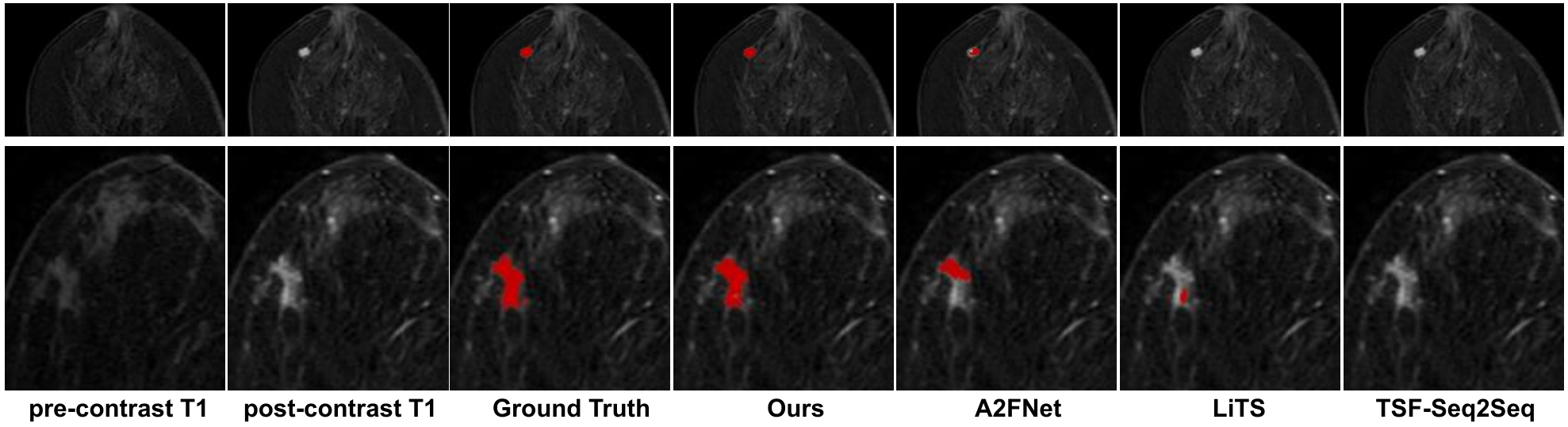}
\vspace{-0.4cm}
\caption{Visualization of breast lesion segmentation. Each row represents a case from our in-house dataset. 
% The first row shows a patient with ductal dilatation behind the right nipple, accompanied by scattered small nodular enhancements, which is very small and difficult to detect. The second patient has a non-mass enhancement with an irregular shape. The third patient has a fibroadenoma with stromal edema, resulting in uneven brightness within the mass.
}
\label{vis_breast}
\vspace{-0.3cm}
\end{figure}

\vspace{-0.3cm}
\section{Experiments}
\vspace{-0.3cm}
\subsection{Dataset}
\vspace{-0.2cm}
% We conduct experiments on two distinct datasets: our in-house breast MRI dataset and the publicly available brain MRI dataset, specifically the BraTS2023 Glioma dataset. 

\noindent \textbf{In-house Breast Dataset:} 
% The breast MRI dataset encompasses a total of 515 cases, with 457 cases allocated for training and 58 for validation. The dataset comprises images acquired from Siemens, GE, and Philips MRI machines, featuring both pre-contrast and post-contrast T1-weighted sequences. Experienced radiologists meticulously annotated lesion masks for each case in our dataset, ensuring high-quality ground truth data.
The breast MRI dataset encompasses a total of 515 cases. The dataset comprises images acquired from Siemens, GE, and Philips MRI machines, featuring both pre-contrast and post-contrast T1-weighted sequences. Experienced radiologists annotated lesion masks for each case in our dataset, ensuring high-quality ground truth.

\noindent \textbf{BraTS2023 Glioma:}
% The BraTS2023 Glioma dataset includes 1126 cases, with 998 cases designated for training and 128 for validation. This comprehensive dataset incorporates T1-weighted, post-contrast T1-weighted, T2-weighted, and Flair sequences. The dataset also provides segmentation masks for tumors. Since our main goal is not multi-sequence fusion, but rather to utilize the contrast info of MRI to adapt to various enhancement patterns, we only select the T1 and T1 contrast sequence from each patient, where T1 contrast serves as main branch and T1 as auxiliary branch. The rest sequences (T2-weighted and Flair sequences) are not used temporarily.
The BraTS2023 Glioma dataset includes 1126 cases. This comprehensive dataset incorporates T1-weighted and post-contrast T1-weighted sequences. The dataset also provides segmentation masks for tumors. 
% Since our main goal is not multi-sequence fusion, but rather to utilize the contrast info of MRI to adapt to various enhancement patterns, we only select the T1 and T1 contrast sequence from each patient, and the rest two sequences are not used temporarily.

% \subsection{Training Details}
% Our training process is conducted using the PyTorch framework, with the Adam optimizer employed for optimization. To accommodate the diverse sizes of the breast MRI and BraTS2023 Glioma datasets, we implement dataset-specific image cropping during the training period. For the BraTS dataset, we utilize a cropping size of 96*96*96, while for our in-house breast dataset, the cropping dimensions are set to 32*80*216. This tailored approach ensures optimal training adaptability to the unique characteristics of each dataset.

% The initial learning rate is 1e-4, and we adopt dice loss function. Other hyper-parameters include: batch size of 8, scheduler using MultiStepLR ($\gamma=0.2$, milestones = [300,500]). To avoid overfitting and enhance the robustness of the model, we employ various data augmentation techniques. These include random flipping, random rotation, random adjustments of brightness and contrast, and the addition of random Gaussian noise.

% Our module can adapt to various encoders and decoders. To maintain comparability with ResUnet, we used encoders from ResUnet architecture in our experiments, consistent with that in \cite{zhang2018road}

\vspace{-0.4cm}
\subsection{Experimental Results}
\vspace{-0.2cm}
Our proposed module can adapt to various encoders and decoders. To maintain comparability with ResUNet, we used encoders from ResUnet architecture in our experiments, consistent with that in \cite{zhang2018road}.

% To assess the effectiveness of our proposed method, we conduct comparative experiments with three recent papers: \textbf{a)} A2FNet~\cite{wang2023a2fseg} proposed an average and adaptive fusion module based on attention mechanism. \textbf{b)} TSF-Seq2Seq~\cite{han2023explainable} utilized a task-specific attention module to synthesize MRI sequences. \textbf{c)} LiTS~\cite{zhang2023multi} integrates complementary information to handle spatial misalignment and segmentation uncertainty. Additionally, we conduct experiments solely on post-contrast images, denoted as Res-UNet-Single. The dataset partition is consistent across these methods, and we utilize Dice Coefficient (DSC) and recall (at pixel level) as evaluation metrics.
To assess the effectiveness of our proposed method, we conduct comparative experiments with three recent papers: A2FNet~\cite{wang2023a2fseg}, TSF-Seq2Seq~\cite{han2023explainable} and LiTS~\cite{zhang2023multi}. We also conduct experiments solely on post-contrast images, denoted as Res-UNet-Single. We utilize Dice Coefficient (DSC) and recall (at pixel level) as evaluation metrics

% \subsubsection{In-house Breast Dataset}
\noindent \textbf{In-house Breast Dataset:} 
Experimental results on our in-house breast MRI dataset are presented in Tab.~\ref{result_breast}. Our proposed approach achieves DSC of 82.55\%, surpassing the performance of all compared methodologies. Additionally, the Recall for our method stands at 82.05\%, demonstrating its effectiveness in accurately capturing relevant information within the breast MRI dataset. The results underscore the superior performance of our proposed method in comparison to state-of-the-art techniques, highlighting its potential for robust and accurate segmentation in breast MRI applications.

\begin{figure}[t!]
\centering
\includegraphics[width=8.5cm]{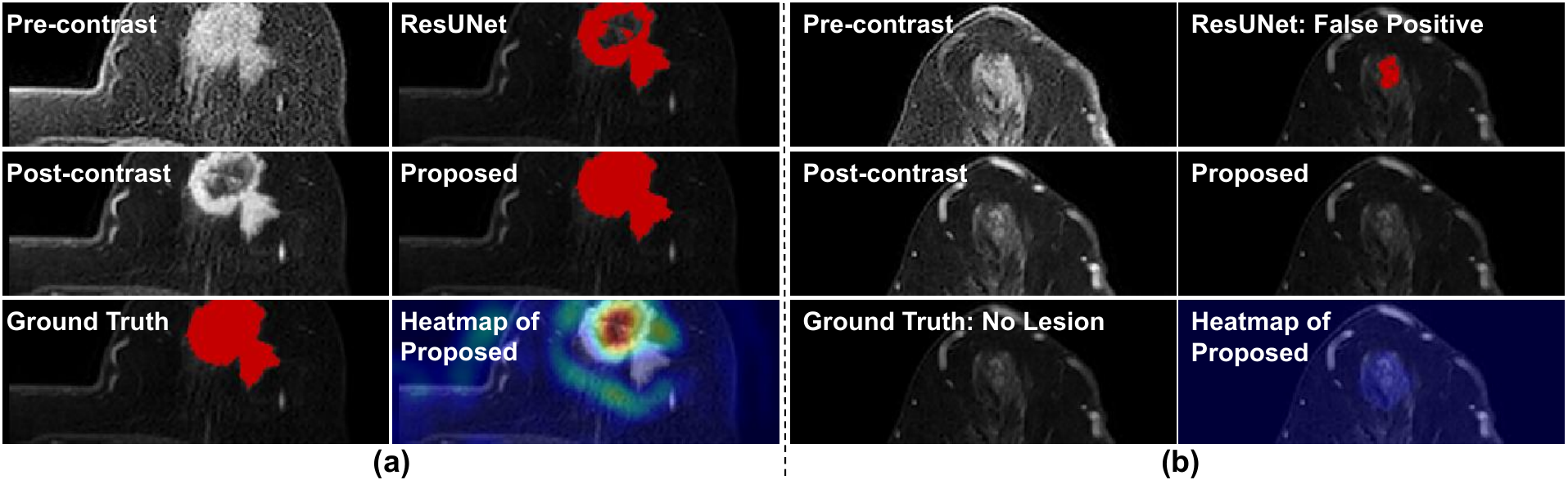}
\vspace{-0.5cm}
\caption{
% Visualization of lesions with atypical enhancement pattern. 
Visualization of atypical enhanced lesion.
% This patient has an irregular mass with lobulated internal structures and marked irregular enhancement after contrast administration, with a central area of low enhancement. Previous models often exhibit significant deficiencies in detecting and outlining the low enhancement area in the center, frequently missing it altogether.
} 
\label{vis_heatmap}
\vspace{-0.1cm}
\end{figure}

Fig.~\ref{vis_breast} shows the segmentation results of different models on breast MRI, each row represents a patient with a lesion segmented out by radiologists. The lesion in the first row is a very small, spot-like enhancement that some models tend to overlook, our model maintains a high detection rate for small lesions. The lesion in the second patient has an irregular shape with additional spot-like local enhancements. Only our model can fully detect the irregular contour of the entire lesion. 
% The third patient has a typical enhancement of the gland, where the entire gland is significantly enhanced, making it easy for the true lesion features to be obscured. The segmentation results from other models exhibit various flaws, such as missing the lesion edges or having gaps in the lesion center, but our model performs well in these cases.

\begin{figure}[t!]
\includegraphics[width=8.5cm]{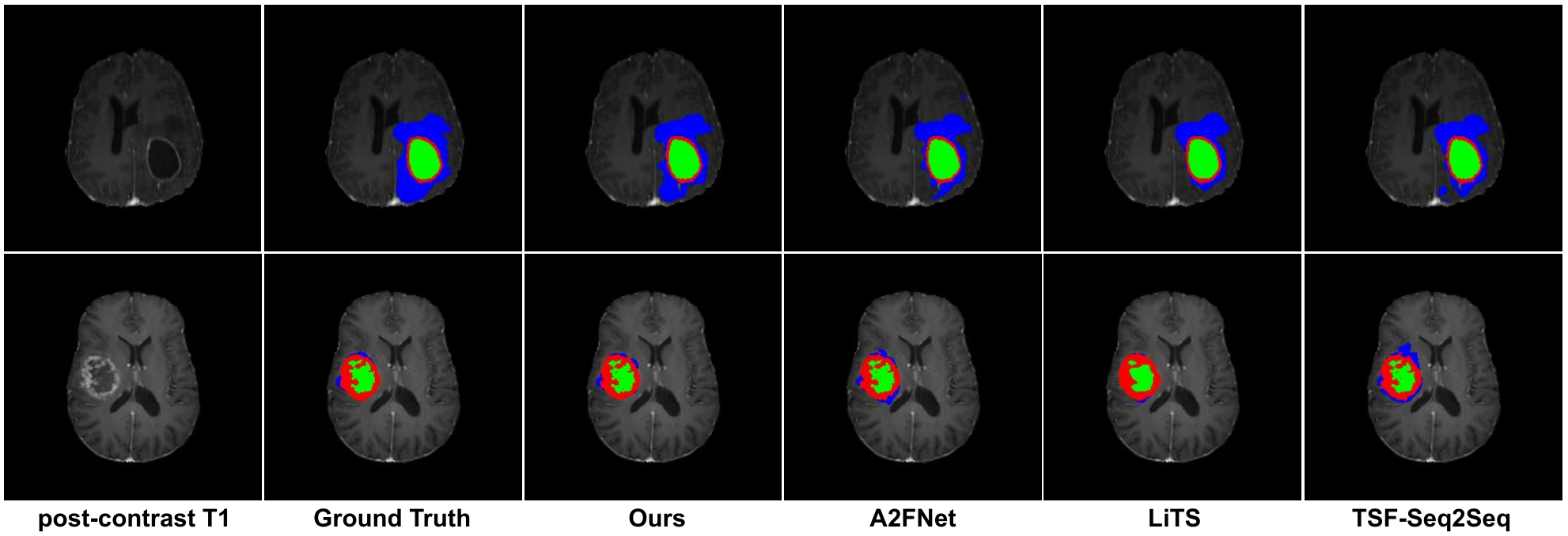}
\vspace{-0.4cm}
\caption{Visualization of brain tumor segmentation. Each row shows a case from the BraTS2023 dataset. WT=red+green+blue; TC=red+green; EN=red.} \label{vis_brain}
\vspace{-0.2cm}
\end{figure}

% We also selected a visualization result to further demonstrate the function of the aux-branch, which is shown in Fig.~\ref{vis_heatmap}. This lesion exhibits significant heterogeneous enhancement after contrast, with the central area being a low-enhancement zone. The second column shows the segmentation results of ResUnet and our proposed model, respectively, which shows that incorporating pre-contrast phase significantly improves the segmentation of the low-contrast area. The bottom right iamge illustrates the heatmap, demonstrating that our model accurately identifies areas that require attention. This holds true even in atypical enhancement areas where the signal intensity decreases after contrast agent injection.

We also select two visualization results to further demonstrate the function of the aux-branch, which is shown in Fig.~\ref{vis_heatmap}. The lesion from patient (a) exhibits significant heterogeneous enhancement after contrast, with the central area being a low-enhancement zone. The second column shows the segmentation results of ResUNet and our proposed model, respectively, which shows that incorporating the pre-contrast phase significantly improves the segmentation of the low-contrast area. The bottom right image illustrates the heatmap, demonstrating that our model accurately identifies areas that require attention. 
% This holds true even in atypical enhancement areas where the signal intensity decreases after contrast agent injection.
Meanwhile, patient (b) is a classic case of glandular enhancement. The ResUNet model's prediction, which does not incorporate pre-contrast information, includes false positives. In contrast, our model mitigates the occurrence of false positives and achieves better performance.

\begin{table}[t!]
\centering
\caption{Comparison among recent methods on BraTS2023 dataset. WT=whole tumor, TC=tumor core, EN=enhancing.}\label{result_brats}
\begin{tabular}{>{\centering\arraybackslash}m{2.8cm} % 使用m{}代替p{}来自动调整列宽
          >{\centering\arraybackslash}m{0.8cm}
          >{\centering\arraybackslash}m{0.8cm}
          >{\centering\arraybackslash}m{0.8cm}
          >{\centering\arraybackslash}m{0.9cm}}
\toprule
\multirow{2}{*}{\textbf{Methodology}} & \multicolumn{4}{c}{\textbf{DSC}} \\
\cmidrule(lr){2-5} % 使用cmidrule代替cline来创建跨列的线条
& \textbf{WT} & \textbf{TC} & \textbf{EN} & \textbf{Mean} \\
\midrule
ResUNet\cite{zhang2018road} & 76.74\% & 86.60\% & 83.52\% & 82.29\% \\ 
% Res-UNet-Concat\cite{zhang2018road} & 79.80\% & 87.92\% & 85.44\% & 84.39\% \\ 
A2FNet\cite{wang2023a2fseg} & 80.96\% & 87.98\% & 84.20\% & 84.38\% \\ 
TSF-Seq2Seq\cite{han2023explainable} & 80.87\% & 87.83\% & 85.73\% & 84.81\% \\ 
LiTS\cite{zhang2023multi} & 81.03\% & 88.34\% & 83.37\% & 84.25\% \\
Ours & \textbf{83.87\%} & \textbf{89.17\%} & \textbf{86.28\%} & \textbf{86.44\%} \\ 
\bottomrule
\end{tabular}
\vspace{-0.5cm}
\end{table}

% \subsubsection{BraTS2023 Glioma Dataset}
% \noindent \textbf{BraTS2023 Glioma Dataset:} 
% In Table \ref{result_brats}, a thorough examination of recent methodologies on the BraTS2023 Glioma dataset reveals the segmentation performance measured through DSC across distinct tumor regions, namely Whole Tumor, Tumor Core, and Enhancing regions. Our approach achieves the highest DSC values across all tumor regions: 83.87\% for WT, 89.17\% for TC, and 86.28\% for EN. The overall mean DSC for our method stands at 86.44\%, outshining all other methods in the comparison. The consistently superior performance across diverse tumor regions signifies the efficacy and reliability of our approach, positioning it as a compelling solution for precise and accurate glioma segmentation in clinical settings.

% Fig.~\ref{vis_brain} illustrates the segmentation results of different models on brain MRI. In the task of brain tumor segmentation, the delineation of the boundary between brain tissue structures and the lesion area is particularly crucial, especially for the tumor edema region (indicated in blue in the figure), which is challenging to distinguish precisely from the surrounding brain tissue due to its subtle features across various imaging sequences. However, compared to other comparative models, our proposed model achieves clearer boundary demarcation in the tumor edema region while maintaining excellent segmentation accuracy in the tumor core area.

\noindent \textbf{BraTS2023 Glioma Dataset:} 
Tab. \ref{result_brats} shows DSC values for Whole Tumor (WT), Tumor Core (TC), and Enhancing (EN) regions on the BraTS2023 Glioma dataset. Our method achieves the highest DSCs: 83.87\% for WT, 89.17\% for TC, and 86.28\% for EN, with an overall mean of 86.44\%. This consistent superiority shows the effectiveness of our approach and its potential for accurate glioma segmentation in clinical settings.
% This consistent superiority highlights the effectiveness of our approach and the potential for accurate glioma segmentation in clinical settings.
% his result highlights the effectiveness and potential of our approach for accurate glioma segmentation in clinical settings.
Fig.~\ref{vis_brain} shows brain MRI segmentation results from different models. Accurate boundary delineation between brain tissue and lesion areas, particularly the challenging tumor edema region (blue in the figure), is crucial. Our model outperforms others by providing clearer edema boundaries while maintaining high core tumor segmentation accuracy.

\vspace{-0.4cm}
\section{Conclusion}
\vspace{-0.3cm}
%Inspired by clinical diagnostic practices of radiologists, we propose a novel architecture that integrates information from both pre-contrast and post-contrast magnetic resonance images. A distinctive feature of this structure is the multi-scale fusion of image information with dynamically adjusted fusion weights. Despite not causing a substantial rise in the parameter count, experiments conducted on both our in-house breast dataset and the BraTS2023 Glioma dataset showcase the superior performance of this network in segmentation tasks. The model enhances the precision of lesion boundaries and refines the detection of specific abnormalities. Moreover, our proposed design exhibits versatility, as it can be seamlessly applied to various existing segmentation backbones.
Inspired by the clinical diagnostic practices of radiologists, we introduce a new method that combines data from both pre- and post-contrast MRI scans. This innovative approach is characterized by its multi-scale fusion of images, which is adeptly modulated by dynamically adjusted fusion weights. This method is pivotal as it enriches the model’s ability to delineate lesion boundaries with heightened precision and to discern subtle abnormalities with greater clarity. Despite not causing a substantial rise in the parameter count, experiments conducted on both our in-house breast and the BraTS2023 Glioma dataset show the superior performance of this network in segmentation tasks. Moreover, our method stands out as a clinically inspired solution, promising to elevate the standards of MRI lesion segmentation in line with the real-world demands of medical diagnostics.

\vspace{-0.4cm}
\section*{Acknowledgments}
\vspace{-0.4cm}
This work was supported by the State Key Lab of General Artificial Intelligence at Peking University and Qualcomm University Research Grant.

% \vfill
% \pagebreak

{\small
% \bibliographystyle{IEEEbib}
% \bibliography{strings,refs_mine}
\bibliographystyle{./IEEEabrv_etc/IEEEtran}
% \bibliography{./IEEEabrv_etc/IEEEabrv,refs_mine} 

}

\end{document}